\documentclass[final]{cvpr}

\usepackage[margin=6pt,font=small,labelfont=bf,labelsep=endash,tableposition=top]{caption}

\usepackage{epsfig}
\usepackage{graphicx}
\usepackage{amsmath}
\usepackage{amssymb, nicefrac}

\usepackage{float}
\usepackage{graphicx}
\usepackage{comment}
\usepackage{amsmath,amssymb} %
\usepackage{color}
\usepackage{xspace}
\usepackage{textcomp}
\usepackage{cite}
\usepackage{mathtools}
\usepackage{bbm}
\usepackage{makecell, verbatim, sidecap}
\usepackage{subfig}
\usepackage{multirow}
\usepackage{url,xspace}
\usepackage{xfrac}

\renewcommand{\vec}[1]{\ensuremath{\pmb{#1}}}
\newcommand{\mat}[1]{\ensuremath{\mathbf{#1}}}
\newcommand{\set}[1]{\ensuremath{\mathscr{#1}}}

\makeatletter
\@tfor\next:=abcdefghijklmnopqrstuvwxyxz\do
{\begingroup\edef\x{\endgroup
		\noexpand\@namedef{v\next}{\noexpand\vec{\next}}%
	}\x}
\@tfor\next:=ABCDEFGHIJKLMNOPQRSTUVWXYZ\do
{\begingroup\edef\x{\endgroup
		\noexpand\@namedef{m\next}{\noexpand\mat{\next}}%
	}\x}
\@tfor\next:=ABCDEFGHIJKLMNOPQRSTUVWXYZ\do
{\begingroup\edef\x{\endgroup
		\noexpand\@namedef{s\next}{\noexpand\set{\next}}%
	}\x}
\makeatother

\def\eg{{\it e.g.}\xspace}
\def\ie{{\it i.e.}\xspace}

\def\R{{\mathbb R}}

\def\softmax{{\rm softmax}}

\DeclarePairedDelimiter{\floor}{\lfloor}{\rfloor}

\def\Ours{{FCPose}\xspace}

\usepackage{xcolor}		
\usepackage[colorlinks = true,
            linkcolor = purple,
            urlcolor  = blue,
            citecolor = cyan,
            anchorcolor = black]{hyperref}

\def\confYear{CVPR 2021}

\begin{document}

\title{\Ours: Fully Convolutional Multi-Person Pose Estimation 
    \\
        with Dynamic Instance-Aware Convolutions\thanks{ 
        \it Appearing in Proc.\  IEEE Conf.\  Computer Vision and Pattern Recognition (CVPR), 2021.
 Content may be slightly different from the final published version.
        }
        }
\author{
{Weian Mao}$^1$,      ~~~~~~~ ~~
Zhi Tian$^1$,       ~~~~~~~ ~~
Xinlong Wang$^1$,   ~~~~~~~ ~~
Chunhua Shen$^{1,2}$\thanks{Corresponding author (email: {\tt chunhua@me.com}).
} \\
[0.152cm]
$^1$ The University of Adelaide, Australia
~~~~~~~~~
$^2$ Monash University, Australia
}

\makeatletter
\let\@oldmaketitle\@maketitle%
\renewcommand{\@maketitle}{\@oldmaketitle%
 \centering
    \includegraphics[width=\linewidth]{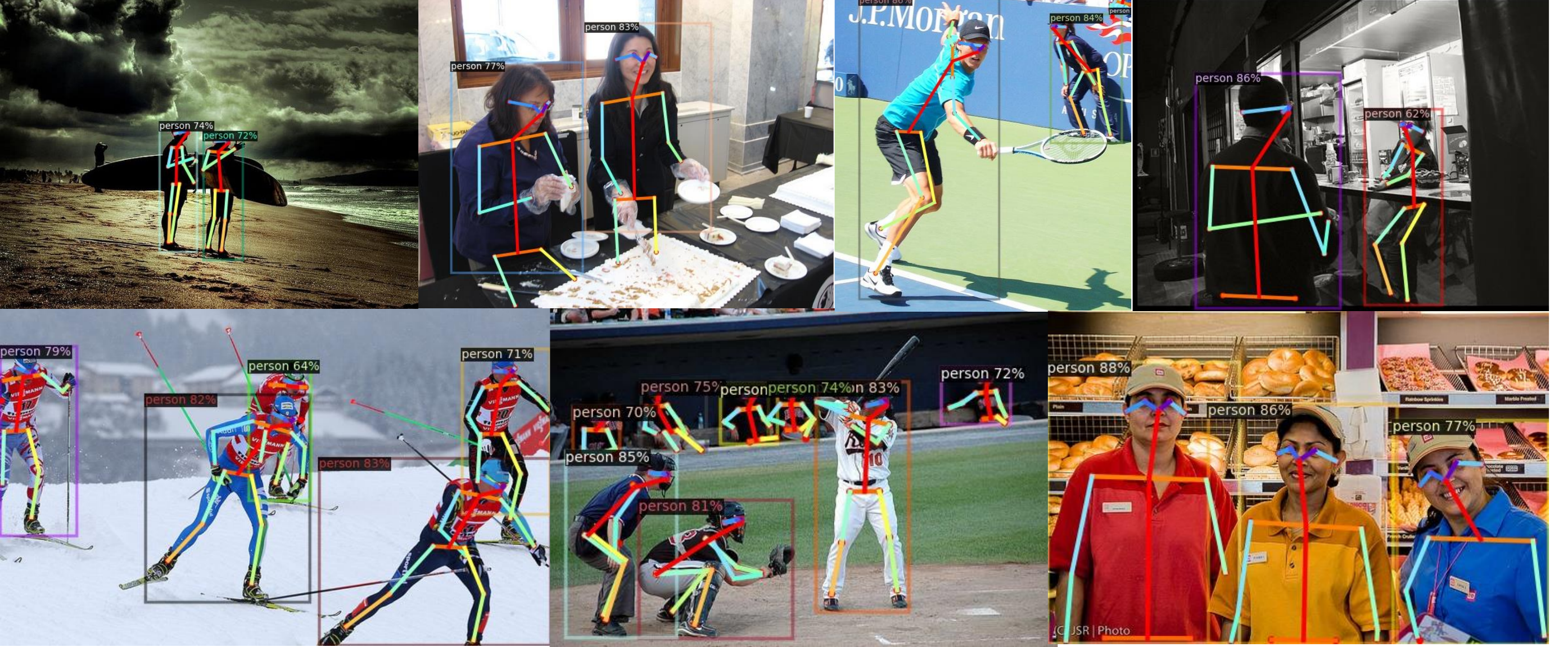}
    \captionof{figure}{
      \textbf{Qualitative results of \Ours.} 
   The results are %
   obtained using 
   the ResNet-101 based \Ours, achieving 65.6\% AP$^{kp}$ on the MS-COCO \texttt{test}-\texttt{dev} split.
    }
    \label{fig:vis_results}
\bigskip}                        %
\makeatother

\maketitle

\begin{abstract}

We propose 
a fully convolutional multi-person pose estimation framework %
using 
dynamic instance-aware convolutions, termed \Ours. Different from %
existing methods, which often require ROI (Region of Interest) operations and/or grouping post-processing, \Ours\ eliminates the ROIs and grouping post-processing with dynamic instance-aware keypoint 
estimation 
heads.  The dynamic keypoint heads are conditioned on each instance (person), and can encode the instance concept  in the dynamically-generated weights of their filters. Moreover, with the strong representation capacity of dynamic convolutions, the keypoint heads in \Ours\ %
are
designed to be very compact, 
resulting  in fast inference and %
making
\Ours\  have almost constant inference time regardless of the number of persons in the image. For example, on the COCO dataset, a real-time version of \Ours\  using the  DLA-34 backbone infers about 4.5$\times$ faster than Mask R-CNN (ResNet-101) 
(41.67  FPS vs.\  9.26 FPS) while achieving %
improved 
performance (64.8\%  AP$^{kp}$ vs.\ 64.3\%  AP$^{kp}$). \Ours\ also 
offers
better speed/accuracy trade-off than other state-of-the-art methods. 
Our
experiment results show that \Ours\ is a simple yet effective multi-person pose estimation framework.
Code is available at:
\def\UrlFont{\tt \color{blue}}
\url{https://git.io/AdelaiDet}

\end{abstract}

\section{Introduction}

\begin{figure}[t]
\small
\centering
\includegraphics[width=\linewidth]{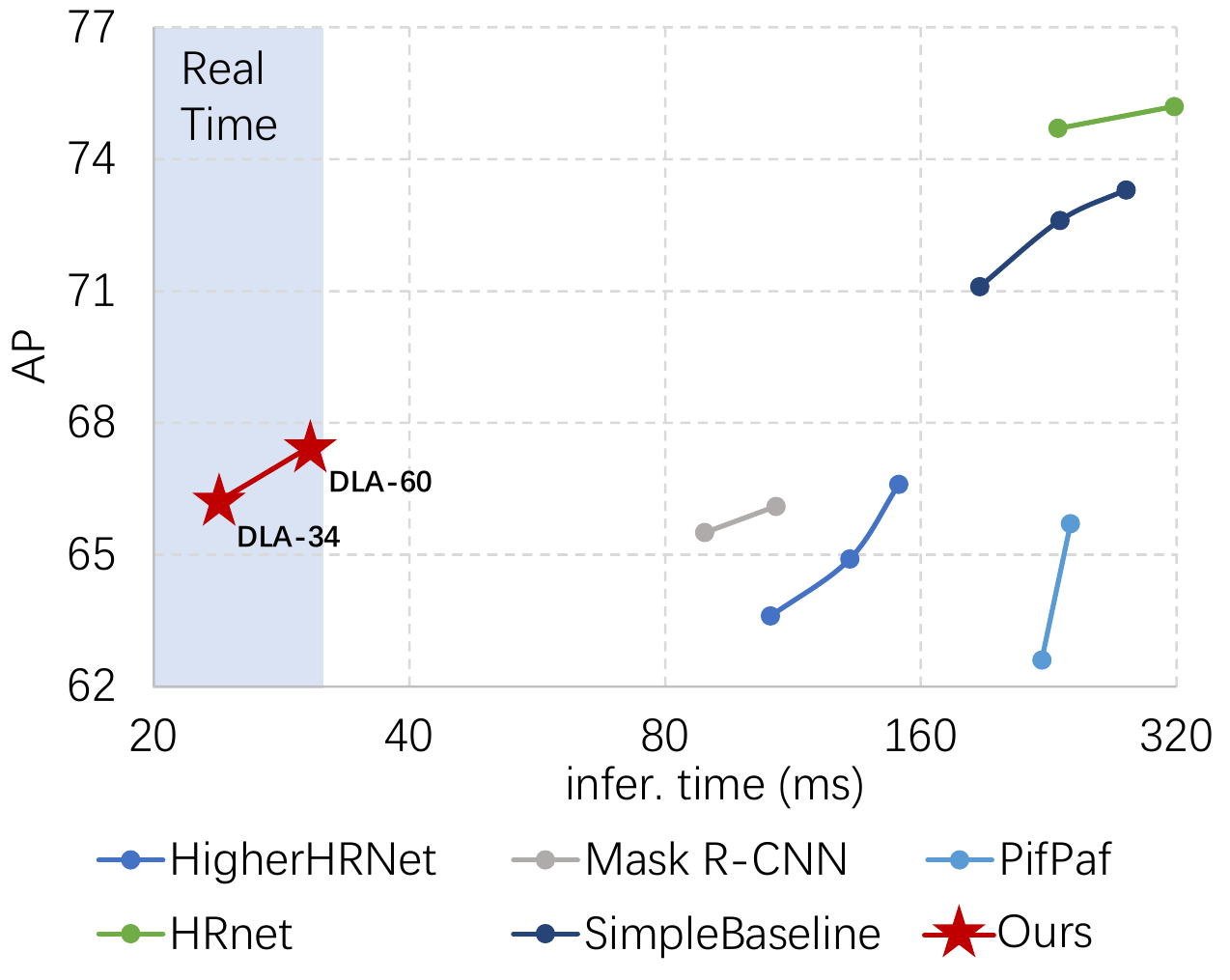}
\caption{\textbf{The trade-off between speed and accuracy}. 
Inference time is measured on a single 1080Ti GPU. %
We report the AP$^{kp}$ on the COCO \texttt{val2017} dataset. 
\Ours\ (DLA-34 backbone \cite{yu2018deep})
is up to $13\times$ faster than previous state-of-the-art methods and achieves real-time speed 
with
competitive 
accuracy.
It is worth noting that in the fast/real-time keypoint detection realm, which is previously dominated by
bottom-up methods, \Ours\  surpasses them both in speed and accuracy. This suggests that \Ours %
serve as 
a new strong baseline for %
real-time keypoint detection.}
\label{fig:tradeoff}
\vspace{-2em}
\end{figure}

Multi-person pose estimation (\textit{a.k.a.},   keypoint detection) aims to obtain the keypoint locations of the persons in an image, which is one of the fundamental computer vision tasks 
with many downstream applications.

The key challenge in multi-person keypoint detection is how to obtain the \textit{instance-level} keypoints. In other words, the detected keypoints need to be grouped according to the instance 
that 
they belong
to. Currently, the mainstream methods tackle this challenge with 
bottom-up or top-down %
approaches. 
Top-down methods~\cite{he2017mask, sun2019deep} first detect each individual instance with a person detector. These detected boxes form the ROIs and an ROI operation is used to crop the person from 
either 
the feature maps or the original image. Next, 
single person keypoint detection 
is performed 
within a ROI for each person, individually. The ROI-based pipeline may come with some drawbacks. First, the ROIs are forwarded separately and thus the convolutional computation cannot be shared. As a result, the inference time of these methods heavily depends on the number of instances in the image, which impedes these methods from being real-time, as shown in our experiments. Second, 
top-down methods usually are not  end-to-end trainable since the ROIs are often obtained from an isolated person detector such as Faster R-CNN~\cite{ren2015faster}. Moreover, the use of the isolated detector also results in significantly longer end-to-end inference time (\ie, from the raw image to the final keypoint results). Third, these ROI-based methods also rely on the localization quality of the ROIs. This may harm the keypoint detection performance if the detector yields 
inaccurate
boxes. On the other hand, bottom-up methods~\cite{cao2018openpose, newell2017associative} do not rely on ROIs. They first detect instance-agnostic keypoints and then employ 
grouping post-processing to obtain the full-body results
for each instance. The processing of assembling the keypoints is usually heuristic and can involves many hyper-parameters, making these methods complicated.

In this work, we propose a new solution %
to 
keypoint detection. Our 
solution is simple yet effective, and is able to avoid the shortcomings of the previous ROI-based or grouping-based methods. \textit{The key idea of our solution is to use the keypoint heads whose convolution filters/weights are dynamically generated}. More specifically, for each instance, we 
dynamically generate a keypoint head. The generated keypoint head is applied to convolutional feature maps in the fashion of fully convolutional networks.
As a result, we are 
able to obtain the keypoint detection results \textit{only for that specific target instance}, as shown in Fig.~\ref{fig:keypoint_head}. This is made possible 
as 
the keypoint head can encode the instance's characteristics in the filters' weights. Thus, this keypoint head can distinguish the instance's keypoints from that of other instances, 
hence \textit{instance-specific convolution filters}. If we consider that the ROI operations in top-down methods are the operations making the model attend to an instance, then in our method, so do the dynamic instance-aware keypoint heads.  This idea eliminates the need for ROIs and the grouping post-processing, thus bypassing the drawbacks mentioned before. Additionally, the dynamically-generated keypoint head is very compact and only has several thousand
coefficients. 
Thus, it can infer very fast, making the overall inference time  almost
remain
the same 
regardless of 
the number of persons in a test image.
This is particularly valuable for real-time applications.

Moreover, 
it is easy to see that the 
localization precision of keypoint detection 
is tightly related to the output resolution of the FCN.
Typically,  
the output resolution of the fully convolutional keypoint heads is %
designed 
to that of the input feature maps (\eg, $\nicefrac{1}{8}$ of the input image), which is not sufficient for keypoint detection. Simply using de-convolutions, as in Mask R-CNN~\cite{he2017mask}, to upsample the outputs 
would inevitably 
result in significantly increased computation overhead. 
Here, we tackle this dilemma of accuracy vs.\ computation complexity  by proffering a new 
keypoint refinement module.  As shown in Table~\ref{table:upsample_method}, 
compared with the baseline, 
our proposed keypoint refinement can dramatically improve the accuracy (56.2\% \textit{vs.}\ 
63.0\%  AP$^{kp}$) with negligible 
computation overhead (69 ms \textit{vs.}\ 71 ms).

We summarize our contributions as follows.

\begin{itemize}
\itemsep -.05112cm

\item We propose an efficient and accurate human pose estimation framework,
termed \textbf{\Ours}, built upon dynamic filters~\cite{jia2016dynamic}. 
For the first time, we demonstrate that an ROI-free and grouping-free end-to-end trainable human pose estimator can achieve even better accuracy and speed, comparing favourably with %
recent
top-down and bottom-up methods.

\item The elimination of ROIs 
enables 
\Ours %
to
be implemented by only convolution operations, resulting in much easier deployment in %
cases such as on mobile devices. Moreover, %
not using 
ROIs also avoids that the keypoint prediction is truncated by the inaccurate detected boxes as in ROI-based frameworks such as Mask R-CNN~\cite{he2017mask} (\textit{e.g.}, see Fig.~\ref{fig:ours_vs_maskrcnn}).

\item 
The core of \Ours is the use of 
the dynamic filters in our keypoint heads.
Dynamically generated filters 
have demonstrated 
strong representation capacities. 
Thus, we only need 
a small number 
of such convolutions for achieving  top  results. 
As a result, the keypoint heads are very compact and thus the overall inference time is fast and almost constant regardless of the number of the instances in the image. We also present a real-time end-to-end keypoint detection models with competitive performance. The trade-off between speed and accuracy is shown in Fig.~\ref{fig:tradeoff}. We believe that \Ours\ can be a new strong baseline for keypoint detection, particularly in the real-time realm.

\end{itemize}

\section{Related Work}

\noindent\textbf{Multi-person pose estimation.} 
Multi-person pose estimation is often solved using either
top-down or bottom-up 
approaches. 
Almost all top-down methods first obtain the boxes of the person instances with an object detector~\cite{he2017mask, xiao2018simple, sun2019deep, papandreou2017towards, huang2017coarse, fang2017rmpe,iqbal2016multi}. Then, the boxes are 
used to crop the image  patches (\ie, ROIs)
 from the 
 input image.
It is expected that the cropped image patch 
should 
include only one person instance.
A single-person estimation method is applied to the cropped image patch 
to attain
keypoint 
locations. 
These methods %
can work well.
The main drawback is the slow inference speed
because they do not share the computation and features with the %
person 
detector. Thus the second-stage pose estimation can
be very slow when the number of person instances in the image is large.
 Instead of cropping the ROIs from the original image and recomputing the features for them, Mask R-CNN~\cite{he2017mask} proposes the ROIAlign operation, which can directly obtain the features of the ROIs from the feature maps of the detector. Thus, it can share the features between the ROIs and the detector, significantly speeding up the inference.

Bottom-up methods~\cite{pishchulin2016deepcut, cheng2018revisiting, cheng2020higherhrnet, cao2017realtime} usually detect 
all the keypoints in an instance-agnostic fashion, and then a grouping post-processing is used to obtain the instance-level keypoints. For example, CMU-Pose~\cite{cao2018openpose} proposes Part Affinity Fields (PAFs) to group the keypoints,.
The authors of \cite{newell2017associative} employ the associative embedding (AE) to assemble them. Compared to %
top-down methods, %
bottom-up methods are often 
faster because it %
computes all the convolutional features once, being fully convolutional models.

Furthermore, 
built upon
the anchor-free detector FCOS~\cite{tian2019fcos}, DirectPose~\cite{tian2019directpose} proposes to directly regress the instance-level keypoints by considering the keypoints as a special bounding-box with more than two corners. DirectPose can also eliminate the ROIs and grouping post-processing. Compared to DirectPose, 
\textit{\Ours\  takes advantage of the dynamic keypoint head and 
achieves
significantly 
better performance.}

\noindent\textbf{Dynamic filters and conditional convolutions.} The core idea of dynamic filter networks~\cite{jia2016dynamic} and CondConv~\cite{yang2019condconv} is to dynamically generate the weights of the convolutions. 
This is different from the traditional convolutional networks, whose weights are fixed once trained.  Since the weights are dynamically-generated and only used once, the model can have strong representation capacity, even with fewer parameters. Moreover, the dynamic filters can be conditioned on each instance in the image, which can make the filters only fire for the target instance. Thus, it can be viewed as a new operation that makes a model attend to the instance, thus replacing the previous ROI operations.
Relevant work employing dynamic convolutions for solving instance segmentation can be found in \cite{tian2020conditional,solov2}.

\section{Our Approach}

\subsection{Overall Architecture}
\begin{figure}
\centering
    \includegraphics[width=1.02\linewidth]{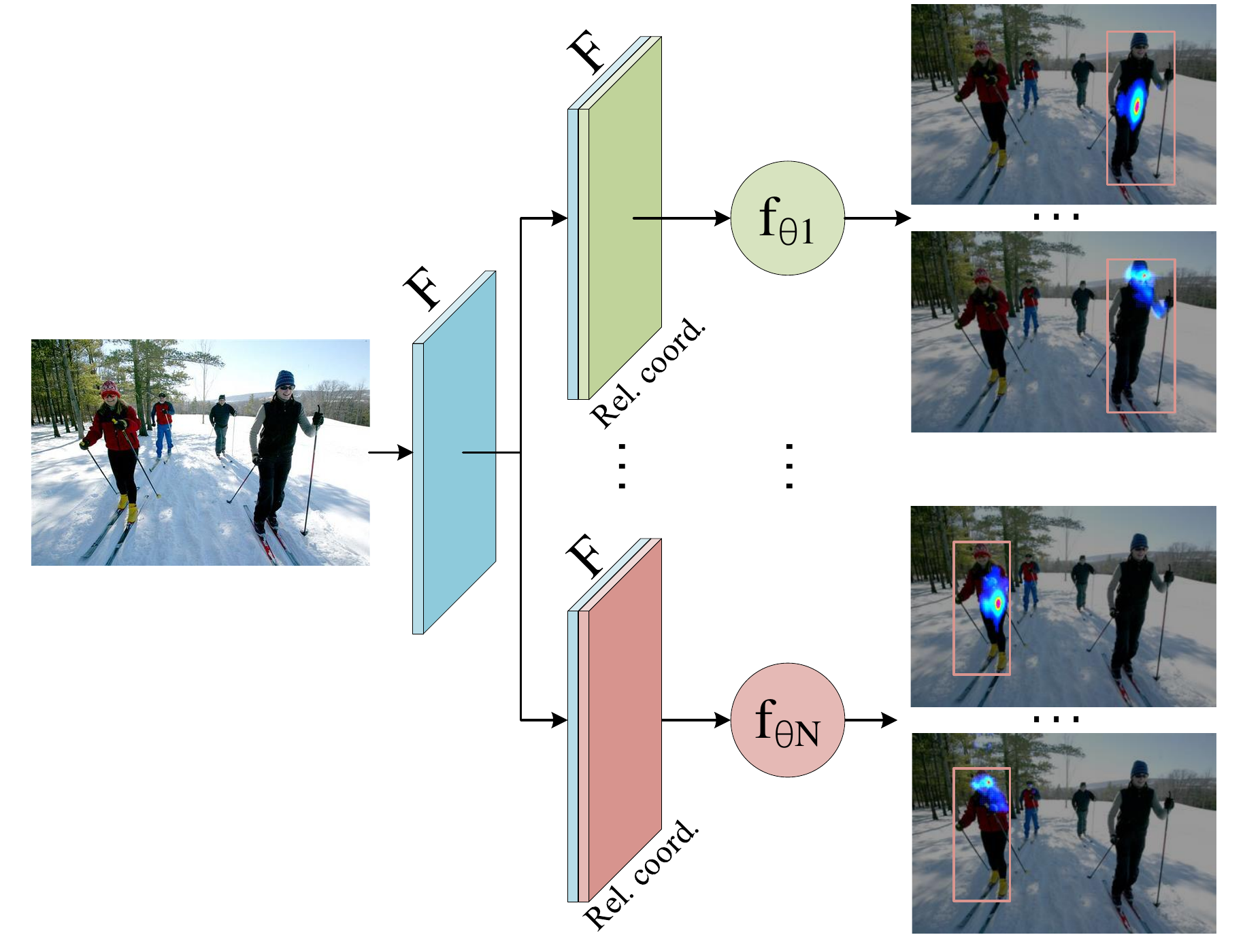}
   \caption{
   \textbf{The core idea of the dynamic keypoint head in \Ours}. $F$ denotes a level of feature maps. ``Rel.\  Coord." means the relative coordinates, denoting the relative offsets from the locations of $F$ to the location where the filters are generated. Refer to the text for details.
   $f_{\theta_i}$ 
   is 
   the dynamically-generated keypoint head for the $i$-th person instance. Note that each person instance has its own keypoint head.}
\label{fig:keypoint_head}
\vspace{-1.0em}
\end{figure}

\begin{figure*}[h]
\centering 
\includegraphics[width=0.986\textwidth]{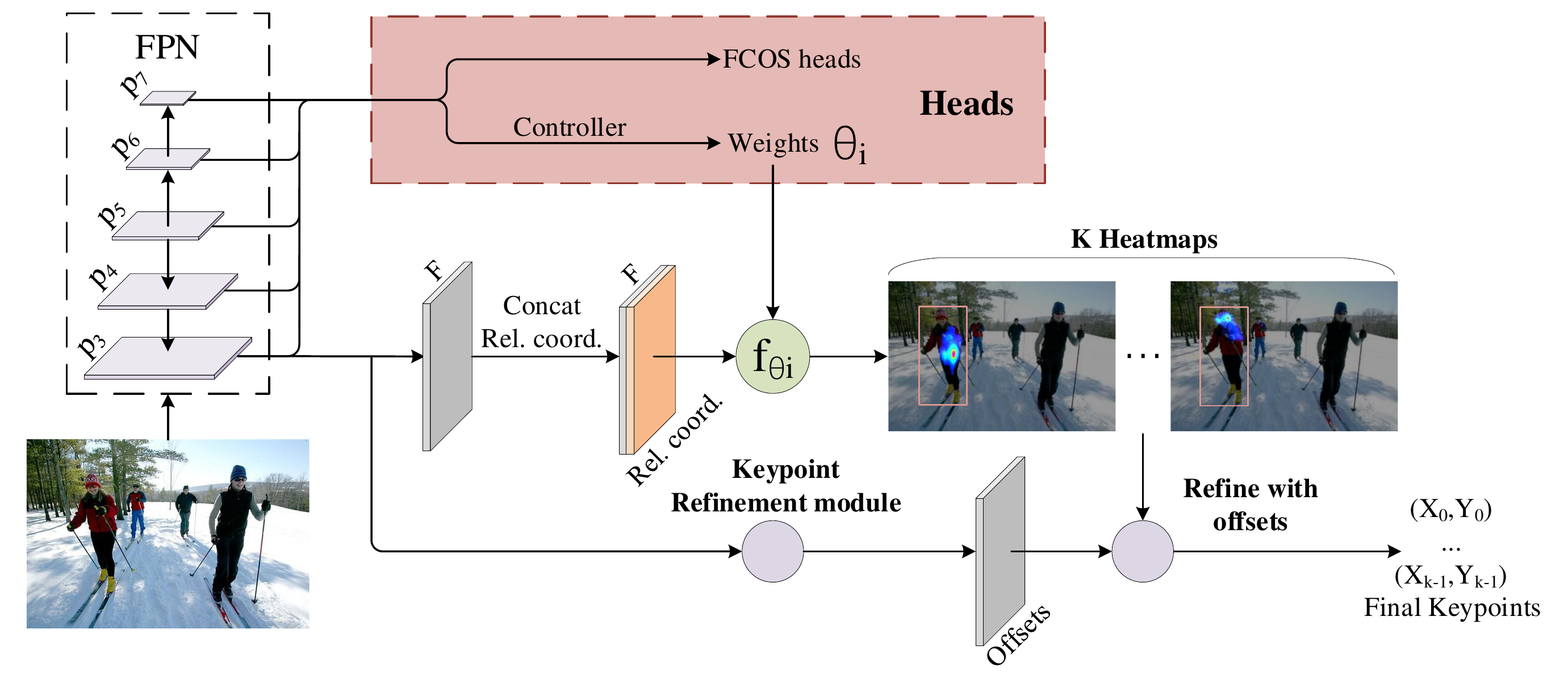}
\caption{
\textbf{The overall framework of \Ours}. \Ours is built upon 
the one-stage object detector FCOS. The controller that generates the weights of the keypoint heads is attached to the FCOS heads. The weights $\theta_i$ generated by the controller is used to
fulfill 
the keypoint head $f$ for the instance $i$. Moreover, 
a keypoint refinement module is introduced to %
predict %
 the offsets from each location of the heatmaps to the ground-truth keypoints. Finally, the coordinates derived from the predicted heatmaps are refined by the offsets predicted by the keypoint refinement module, resulting in the final keypoint results. ``Rel.\  coord." is a map of the relative coordinates from all the locations of the feature maps $F$ to the location where the weights are generated. The relative coordinate map is concatenated to $F$ as the input to the keypoint head.}
\label{fig:framework}
\end{figure*}

In previous works, multi-person keypoint detection is often %
solved as 
per-pixel heatmap prediction with FCNs~\cite{long2015fully}. Since the vanilla FCNs cannot produce instance-aware keypoints, which poses the key challenge in multi-person keypoint detection. As mentioned before, some of methods use an ROI to crop the person of interest and then reduce the multi-person keypoint detection to the single-person one. Formally, let $G \in \R^{h \times w \times c}$ be the features of an ROI, and $f_\theta$ be the keypoint head, where $\theta$ is the learnable network weights. The predicted heatmaps $H \in \R^{h \times w \times K}$ are
\begin{equation}
    H = f_\theta(G).
\end{equation}
Note that $K$ being 17 on COCO is the number of keypoints for an instance. Then, the final keypoint coordinates can be obtained by finding the peak on each channel of the heatmaps. \textit{The ROI operation is the core operation making the model attend to an instance.} In this work, we propose to employ the instance-aware keypoint heads to make the model attend to an instance. For each instance, %
$i$, a new set of weights $\theta_i$ of the keypoint head will be generated. The keypoint head with weights $\theta_i$ is applied to full-image feature maps.

Formally, let $F \in \R^{H \times W \times 32}$ be a level of feature maps, which are generated by applying a few conv.\ layers to the FPN output feature maps~\cite{lin2017feature} and have the same resolution of $P_3$ in the FPN, as shown in Fig.~\ref{fig:framework}. For the instance $i$, the predicted heatmaps $H \in \R^{H \times W \times K}$ are
\begin{equation}
    H = f_{\theta_i}(F).
\end{equation}
Note that $F$ is the full-image feature maps without any cropping operations. The filters' weights $\theta_i$ are conditioned on the features of the instance $i$, and thus it can encode the characteristics of the target instance. This makes it possible that the keypoint head only fires at the keypoints of the instance, as shown in Fig.~\ref{fig:keypoint_head}.

In this work, we use FCOS~\cite{tian2019fcos} to generate the dynamic weights $\theta$ for each instance. To this end, we add a new output branch to the box regression branch of FCOS (\ie, the controller shown in Fig.~\ref{fig:framework}). Recall that in FCOS, each location (if considered positive) on the feature maps is associated to an instance. Thus, if a location is associated to the instance $i$, the controller can generate the weights $\theta_i$ that are used to detect the keypoints of the instance $i$. In this paper, the controller is a single convolutional layer with kernel size $1\times1$. The number of outputs of the layer is equal to the number of the weights in the keypoint head (\ie, the cardinality of $\theta_i$). In this paper, the keypoint head has $3\times$ conv.\ layer with channel 32 and kernel size $1\times1$, followed by ReLU, as well as a $K$-channel final prediction layer (one channel for one keypoint), which has 2, 737 weights in total. Therefore, the final prediction layer of the controller has 2, 737 output channels. It is worth noting that our keypoint head is much more compact than other top-down methods, for example, Mask R-CNN has $8\times$ conv.\ layers with channel 512 and kernel size $3\times3$ as well as deconv.\ layers in the keypoint head. The very compact keypoint head makes our method take negligible inference time on the keypoint head and thus the overall inference time is almost constant regardless of the number of the instances in the image, as shown in Fig.~\ref{fig:ninst_cropped}.

Moreover, provided that the model is predicting the instance with the filters generated at the location $(x, y)$, it is obvious that the locations on $F$ that are far from $(x, y)$ are less likely to be the keypoints for the instance. Thus, inspired by CoordConv~\cite{liu2018intriguing}, we append the relative offsets to the feature maps $F$, as shown in Fig.~\ref{fig:keypoint_head} and Fig.~\ref{fig:framework}, which denote the distances from each location of $F$ to the location $(x, y)$. The keypoint head takes as input the augmented feature maps. This improves the performance remarkably.

\subsection{Keypoint Refinement with Regression}
As mentioned before, the FPN feature maps $P_3$ is used to generate the heatmaps, and thus the resolution of the heatmaps is $\sfrac{1}{8}$ resolution of the input image. Since keypoint detection requires high localization precision, the $\sfrac{1}{8}$ resolution is not sufficient for keypoint detection. In most of previous methods, an upsampling operation such as deconvolution is often used to upsample the heatmaps. However, upsampling the heatmaps comes with high computational overheads in \Ours. Specifically, in \Ours, for an image, we output $N$ heatmaps with channel $K$, height $H$, and width $W$, where $K$ is the number of keypoints of an instance and $N$ is the number of the instances in the image. These heatmaps will occupy $N \times K \times H \times W$ memory footprint. If we upsample the heatmaps by 8 times, the memory footprint will be increased by 64 times. Also, this will result in much longer computational time.

Here, we address this issue by introducing a regression-based keypoint refinement module. As shown in Fig.~\ref{fig:framework}, the keypoint refinement module is also applied to the FPN level $P_3$, which is a single conv.\ layer with output channel $2K$ (\ie, 34 on COCO). Let $O \in \R^{H \times W \times 2K}$ be the output feature maps of this module. $O_{i, j} = (\Delta x, \Delta y)$ predicts the offsets from the location $(i, j)$ to the nearest ground-truth keypoint. As a result, for a keypoint, if its heatmap's peak is at $(i, j)$, the final coordinates of the keypoint will be $(i + \Delta x, j + \Delta y)$. Experiments show that the refinement module can greatly improve the keypoint detection performance with negligible computational overheads. Note that in our experiments, all instances share the same keypoint refinement module. Although it is possible to dynamically generate the module and make each instance have its own one, we empirically find that using one shared keypoint refinement module is sufficient.

\subsection{Training Targets and Loss Functions}
\paragraph{Training targets.} \Ours\ is built on the detector FCOS. First, we use the same processing to associate each location on the feature maps with an instance or label the location negative. The classification and box regression training targets of each location are computed as in FCOS. As mentioned before, a location is also required to generate the keypoint head's filters for the associated instance. The generated filters are not explicitly supervised. Instead, we supervise the heatmaps predicted by the keypoint head with the filters, which implicitly supervise the generated filters. Except for the classification outputs, all the other outputs are only supervised at the positive locations. In FCOS, for each batch of images (on the same GPU), we might have up to $\sim 500$ positive locations. If all these locations are used to generate the filters, it will come with high computational overheads. Therefore, for each batch, we only sample at most $M=50$ positive locations to generate filters. The $M$ locations are averaged over all the ground-truth instances. For each instance, the positive locations with high confidence will be chosen, and the rest of positive locations will be discarded in the keypoint loss computation.

\paragraph{Loss functions.} The loss functions of \Ours\ consist of three parts. The first part is the original losses of FCOS, which are kept as they are. We refer readers to the paper of FCOS~\cite{tian2019fcos} for the details. The second part is the loss function for the heatmap learning. As mentioned before, one heatmap only predicts one keypoint. Therefore, we can use one-hot training target for the heatmap, and the cross entropy (CE) loss with softmax is used as the loss function. To be specific, assume a ground-truth keypoint's coordinates are $(x^*, y^*)$, and the heatmap's resolution is $\sfrac{1}{8}$ resolution of the input image. Then, for this keypoint, the location $(\floor{\frac{x - 4}{8}}, \floor{\frac{y - 4}{8}})$ on its ground-truth heatmap will be set 1 and other locations will be zeros. Let $H_i^* \in \R^{H \times W}$ be the ground-truth heatmap for the keypoint. The loss function can be formulated as
\begin{equation}
    L_{heatmap} = {\rm CrossEntropy}(\softmax(H_i), H^*),
\end{equation}
where $H_i \in \R^{H \times W}$ is the heatmap predicted by the dynamic keypoint head for this keypoint. Here, both $H_i$ and $H^*_i$ are flatten to a vector, and the cross entropy and softmax are applied to each vector. Finally, for the keypoint offset regression, the mean square error (MSE) is used to compute the difference between the predicted offsets and the ground-truth ones. The overall loss function is the summation of these loss functions. Formally, we have
\begin{equation}
    L_{overall} = L_{fcos} + \alpha L_{heatmap} + \beta L_{reg},
\end{equation}
where $\alpha$ and $\beta$ are the loss weights, respectively.

\subsection{Inference}
Given an input image $I$, \Ours\ first forwards the image through the network and obtain the network outputs. Following FCOS, the locations with the classification score greater than $0.05$ are chosen as positive locations. One positive location corresponds to one predicted instance. Next, for each location, we compute the generated filters and apply the filters to the feature maps $F$ (as shown in Fig.~\ref{fig:framework}) to obtain the keypoint heatmaps of the instance associated with the location. For each heatmap, we find the coordinates of its peak. Then, we refine the coordinates of the peak by the offsets of the keypoint regression module, and obtain the resulting keypoint coordinates. Finally, non-maximum suppression (NMS) is used to remove the duplicates.

\section{Experiments}
We train and evaluate \Ours on the COCO 2017 Keypoint Detection benchmark~\cite{lin2014microsoft}, which has $57K$ images for training, $5K$ images for validation, and $20K$ images for testing. The dataset includes more than $250K$ person instances with 17 annotated keypoints per person. The Average Precision (AP) based on Object Keypoint Similarity (OKS) is used as the evaluation metric. The ablation studies are evaluated on the \texttt{val2017} split. Our main results are reported on the \texttt{test}-\texttt{dev} split.

\noindent\textbf{Implementation details.} We implement \Ours using \texttt{Detecton2}~\cite{wu2019detectron2}. The models are trained with stochastic gradient descent (SGD) over 8 GPUs. Unless specified, all the experiments use the following training details. Following FCOS~\cite{tian2019fcos}, ResNet-50~\cite{he2016deep} with FPNs~\cite{lin2017feature} is used as the feature extractor. The weights pre-trained on ImageNet are used to initialize the backbone ResNet-50. The newly added layers are initialized with the method in \cite{he2015delving}. The learning rate is initially set to $0.01$, and it is reduced by a factor of 10 at iteration $60K$ and $80K$ in the $1\times$ training schedule (\ie, $90K$ iterations), or at $180K$ and $240K$ in the $3\times$ training schedule (\ie, $270K$ iterations). The weight decay, batch size, and momentum are set as 0.0001, 16, and 0.9, respectively. For data augmentation, we apply random crop $[0.4, 1.0]$ (relative range), random flip, and random resizing (the short size of the image is sampled from $[320, 800]$). For inference, we only use single scale of the image. The shorter side of the image is resized to 800 and the longer side is resized to less than or equal to 1333. All the inference time is measured on a single 1080 Ti GPU.

\subsection{Ablation Experiments}
\subsubsection{Architecture of the Dynamic Keypoint Head}
\begin{table}[t]
    \small
    \centering
	\begin{tabular}{c|c|c|cc|cc}
		\# channels & time & AP$^{kp}$ & AP$^{kp}_{50}$ & AP$^{kp}_{75}$ & AP$^{kp}_{M}$ & AP$^{kp}_{L}$ \\
		\Xhline{2\arrayrulewidth}
		16 & \textbf{71}  & 62.8 & 85.7 & 68.6 & 59.0 & 69.9  \\
		32 & \textbf{71}  & \textbf{63.0} & \textbf{85.9} & \textbf{68.9} & \textbf{59.1} & \textbf{70.3} \\
		64 & \textbf{71}  & 62.6 & \textbf{85.9} & 68.3 & 58.8 & 69.9 \\
	\end{tabular}
	\vspace{0.5em}
	\caption{The effect of the number of channels of the input feature maps to the keypoint head (\ie, the feature maps $F$ in the text). ``time": the total inference time per image in milliseconds.}
	\label{table:keypoint_base_channel}
	\vspace{-1.5em}
\end{table}

\begin{table}[t]
    \small
    \centering
	\label{table: the number of convolution layer}
		\begin{tabular}{c|c|c|cc|cc}
			depth & time & AP$^{kp}$ & AP$^{kp}_{50}$ & AP$^{kp}_{75}$ & AP$^{kp}_{M}$ & AP$^{kp}_{L}$ \\
			\Xhline{2\arrayrulewidth}
			2  & \textbf{69} & 62.8 & \textbf{86.0} & 68.7 & 59.0 & 70.0 \\
			3  & 71 & \textbf{63.0} & 85.9 & \textbf{68.9} & \textbf{59.1} & \textbf{70.3}  \\
			4  & 72 & 62.6 & 85.4 & 68.5 & 58.9 & 69.8 \\
		\end{tabular}
	\vspace{0.5em}
	\caption{Varying the number of the layers in the dynamic keypoint head (\ie, depth). ``time": the total inference time per image in milliseconds.}\label{table:num_layers}
	\vspace{-2.5em}
\end{table}

\begin{table}[t]
    \small
    \centering
	\begin{tabular}{c|c|c|cc|cc}
		width & time & AP$^{kp}$ & AP$^{kp}_{50}$ & AP$^{kp}_{75}$ & AP$^{kp}_{M}$ & AP$^{kp}_{L}$ \\
		\Xhline{2\arrayrulewidth}
		16  & \textbf{70}  & 62.5  & 85.5 & 68.0 & 58.3 & 70.1 \\
		32  & 71  & \textbf{63.0}  & \textbf{85.9} & \textbf{68.9} & \textbf{59.1} & \textbf{70.3} \\
		64  & 72  & 62.6  & 85.5 & 68.4 & \textbf{59.1} & 69.5 \\
	\end{tabular}
	\vspace{0.5em}
	\caption{The effect of the number of channels in the dynamic keypoint head (\ie, width). ``time": the total inference time per image in milliseconds.}\label{table:num_channels}
	\vspace{-1.5em}
\end{table}

Here, we study the effect of the architecture of the dynamic keypoint head on the final keypoint detection. Specifically, we conduct experiments by varying the number of channels of the input feature maps, the number of channels of the keypoint head, and the number of conv.\ layers in the keypoint head.

First, we attempt to change the number of channels of the input feature maps (\ie, $F$ mentioned before) of the keypoint head. As shown in Table~\ref{table:keypoint_base_channel}, 16-, 32- and 64-channel input feature maps have roughly the same performance. Among, using 32 channels achieves the best performance. Moreover, we change the number of conv.\ layers in the keypoint head. As shown in Table~\ref{table:num_layers}, the number of conv.\ layers does not significantly affect the final performance but using 3 conv.\  layers is the best. Finally, Table~\ref{table:num_channels} shows the performance is also insensitive to the number of channels of the keypoint head. We use 32 channels in the keypoint head in other experiments since it has the best performance.

\subsubsection{Keypoint Refinement Module}

\begin{table}[t]
    \small
    \centering
    \setlength{\tabcolsep}{5.5pt}
	\begin{tabular}{ l |c|c|cc|cc}
		& time & AP$^{kp}$ & AP$^{kp}_{50}$ & AP$^{kp}_{75}$ & AP$^{kp}_{M}$ & AP$^{kp}_{L}$ \\
		\Xhline{2\arrayrulewidth}
		none & \textbf{69} & 56.2 & 83.6 & 60.8 & 49.8 & 66.3 \\
		deconv. & 135 & 60.1 & 84.8 & 65.7 & 56.4 & 67.3 \\
		proposed & 71 & \textbf{63.0} & \textbf{85.9} & \textbf{68.9} & \textbf{59.1} & \textbf{70.3} \\
	\end{tabular}
	\vspace{0.5em}
	\caption{Comparison of
	various 
	upsampling methods on the COCO \texttt{val2017} split. ``none": %
	no upsamling methods used. ``deconv.": using deconvolutions to upsample the heatmaps. ``proposed": using the proposed keypoint refinement module. As shown
	here, the proposed module %
	achieves 
	much better performance while keeping almost the same inference time as the 
	baseline 
	model without using  any upsampling methods.}
	\label{table:upsample_method}
	\vspace{-1.5em}
\end{table}

\begin{table}[t]
    \small
    \centering
	\begin{tabular}{c|c|c|cc|cc}
		shared & time & AP$^{kp}$ & AP$^{kp}_{50}$ & AP$^{kp}_{75}$ & AP$^{kp}_{M}$ & AP$^{kp}_{L}$ \\
		\Xhline{2\arrayrulewidth}
		 & 71 & 62.7 & 85.2 & \textbf{68.9} & 58.8 & \textbf{70.3} \\
		\checkmark & \textbf{71} & \textbf{63.0} & \textbf{85.9} & \textbf{68.9} & \textbf{59.1} & \textbf{70.3} \\
	\end{tabular}
	\vspace{0.5em}
	\caption{Share the keypoint refinement module between instances or not. As shown in the table, both can have the similar performance. We use the shared one in other experiments due to the slightly better performance.}
	\label{table:shared_keypoint_refinement}
	\vspace{-2.5em}
\end{table}

As mentioned before, the proposed keypoint refinement module can largely improve the localization precision while without introducing large computational overheads. We confirm this in this section.

First, if none of the upsampling methods is used, \Ours\ can only localize the keypoints with the heatmaps that are $\sfrac{1}{8}$ resolution of the input image. Unsurprisingly, this has low keypoint detection performance (56.22\% AP$^{kp}$), as shown in Table~\ref{table:upsample_method}. Most of previous methods such as Mask R-CNN employ deconvolutions to improve the resolution of the heatmaps. However, as mentioned before, deconvolutions will significantly increase the computational overheads. As shown in Table~\ref{table:upsample_method}, using deconvolutions in \Ours\ increases the inference time from 69 ms to 135 ms per image with only 4 points better performance. In contrast, the proposed keypoint refinement module can achieve even better performance than deconvolutions (63.03\% AP$^{kp}$) while keeping almost the same computational time as the model without any upsampling.

Additionally, as mentioned before, we share the keypoint refinement module between all the instances. In principle, it is more reasonable that each instance has its own keypoint refinement module. This is possible by using the dynamic filter technique to generate the keypoint refinement module. However, in Table~\ref{table:shared_keypoint_refinement}, we empirically show that the shared keypoint refinement module can %
achieve 
slightly better performance (62.7\% \textit{vs.}\  63.0\% AP$^{kp}$), and thus the shared one is used in all the other experiments. We conjecture that this is because that the situation where multiple persons crowd together is relatively rare, and thus the unshared one does not %
show
remarkable superiority.

\subsection{Comparisons with State-of-the-art Methods}
\begin{figure}[t!]
\centering
\includegraphics[width=\linewidth]{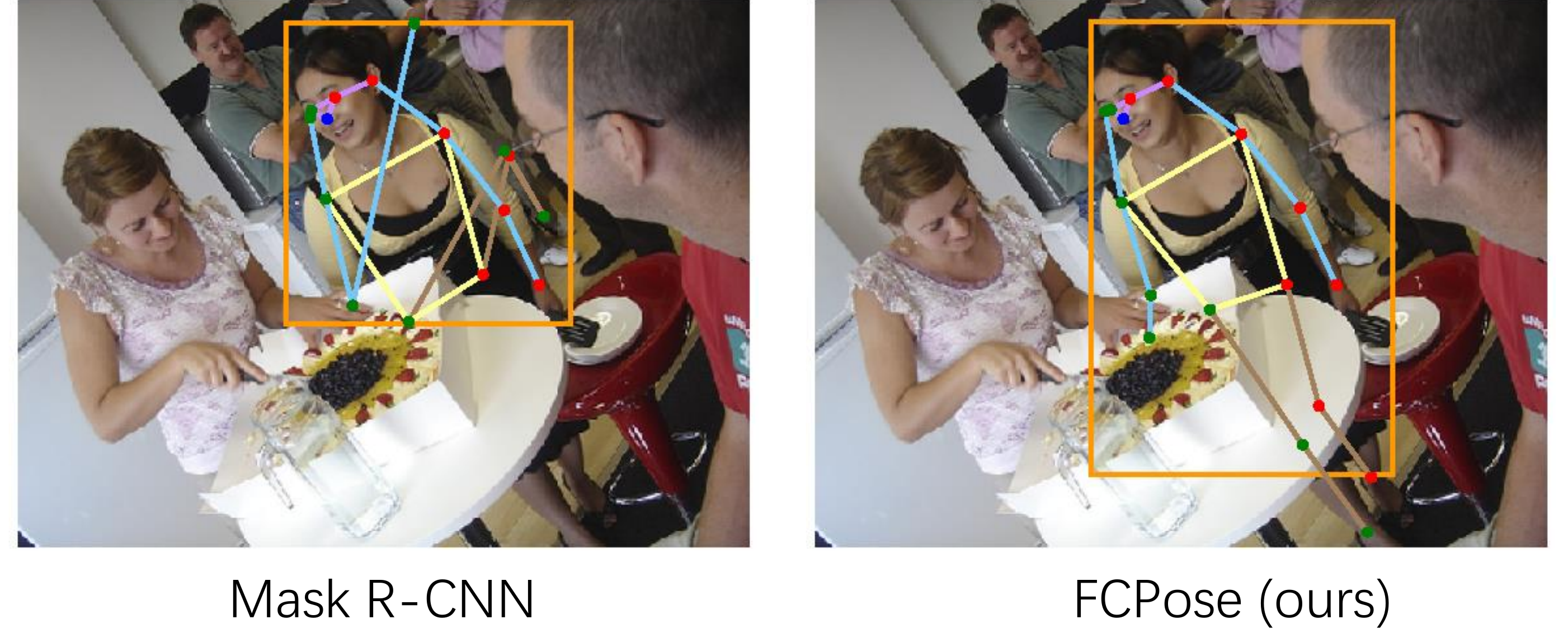}
\caption{Comparison of \Ours and the ROI-based Mask R-CNN. As shown in the figure, Mask R-CNN %
misses some keypoints if the box predicted by the detector is not accurate. In contrast, \Ours can still detect these keypoints since it does not rely on the box.}
\label{fig:ours_vs_maskrcnn}
\end{figure}

\begin{table*}[t!]
    \small
    \centering
    \setlength{\tabcolsep}{5pt}
	\begin{tabular}{ r | l |c|c|c|cc|cc}
		method & backbone & input  size & infer.\  time (ms) & AP$^{kp}$ (\%) & AP$^{kp}_{50}$ & AP$^{kp}_{75}$ & AP$^{kp}_{M}$ & AP$^{kp}_{L}$ \\
		\Xhline{2\arrayrulewidth}
		\multicolumn{8}{c}{\textbf{Top-down methods}} \\
		\hline
		DirectPose~\cite{tian2019directpose} & ResNet-50 & 800 & \textbf{74} & 62.2 & 86.4 & 68.2 &  56.7 & 69.8 \\
		Mask R-CNN~\cite{he2017mask} & ResNet-50 & 800 & - & 62.7 & 87.0 & 68.4 & 57.4 & 71.1 \\
		Mask R-CNN$^*$ & ResNet-50 & 800 & \textbf{89} & 63.9 & 87.7 & 69.9 & 59.7 & 71.5 \\
		Mask R-CNN$^*$ & ResNet-101 & 800 & 108 & 64.3 & 88.2 & 70.6 & 60.1 & 71.9 \\
		G-RMI~\cite{papandreou2017towards} & ResNet-101 & 800 & - & 64.9 & 85.5 & 71.3 & 62.3 & 70.0 \\
		CPN~\cite{chen2018cascaded}        & ResNet-Inc.  & 384$\times$288 & 282 & 72.1 & 91.4 & 80.0 & 68.7 & 77.2 \\
        RSN$^\dag$~\cite{cai2020learning} & RSN-50 & 256$\times$192 & - & 72.5 & 93.0 & 81.3 & 69.9 & 76.5 \\
        SimpleBaseline$^\dag$~\cite{xiao2018simple} & ResNet-152 & 384$\times$288 & 430 & 73.7 & 91.9 & 81.1 & 70.3 & 80.0 \\
        HRNet$^\dag$~\cite{sun2019deep} & HRNet-W32 & 384$\times$288 & 337 & 74.9 & 92.5 & 82.8 & 71.3 & 80.9 \\
		HRNet$^\dag$  & HRNet-W48 & 384$\times$288 & 488 & \textbf{75.5} & \textbf{92.5} & \textbf{83.3} & \textbf{71.9} & \textbf{81.5} \\
		\hline
		\multicolumn{8}{c}{\textbf{Bottom-up methods}} \\
		\hline
		CMU-Pose~\cite{cao2018openpose} & VGG-19~\cite{simonyan2014very} & - & \textbf{74} & 64.2 & 86.2 & 70.1 & 61.0 & 68.8 \\
		AE~\cite{newell2017associative} & HourGlass~\cite{newell2016stacked} & 512 & - & 56.6 & 81.8 & 61.8 & 49.8 & 67.0 \\
		MultiPoseNet$^{\ddagger}$~\cite{kocabas2018multiposenet} & ResNet & 800 & 43 & 69.6 & 86.3 & 76.6 & 65.0 & 76.3 \\
		HigherHRNet$^{\dag\ddagger}$~\cite{cheng2020higherhrnet} & HRNet-W48 & 640 & 1153 & \textbf{70.5} & \textbf{89.3} & \textbf{77.2} & \textbf{66.6} & \textbf{75.8} \\
		HigherHRNet$^\dag$ & HRNet-W48 & 640 & 579 & 68.4 & 88.2 & 75.1 & 64.4 & 74.2 \\
		HigherHRNet$^\dag$ & HRNet-W32 & 512 & 400 & 66.4 & 87.5 & 72.8 & 61.2 & 74.2 \\
		HigherHRNet        & HRNet-W32 & 640 & 128 & 64.7 & 86.9 & 71.0 & 60.2 & 71.2 \\
		\hline
		\multicolumn{8}{c}{\textbf{Our methods}} \\
		\hline
		\textbf{\Ours} & ResNet-50 & 800 & \textbf{68} & 64.3 & 87.3 & 71.0 & 61.6 & 70.5 \\
		\textbf{\Ours} & ResNet-101 & 800 & 93 & \textbf{65.6} & \textbf{87.9} & \textbf{72.6} & \textbf{62.1} & \textbf{72.3} \\
	\end{tabular}
	\vspace{0.5em}
	\caption{Comparisons with recent state-of-the-art methods. $^\dag$ and $^\ddagger$ denote flipping and multi-sacle testing, respectively. We measure the inference time of other methods on the same hardware if possible. Mask R-CNN$^*$ are the results from \texttt{Detectron2}~\cite{wu2019detectron2}, which are better than the original results reported in the Mask R-CNN paper~\cite{he2017mask}.}
	\label{table:comparisons_with_sota}
	\vspace{-2em}
\end{table*}

\begin{figure}[t!]
\small
\centering
\includegraphics[width=\linewidth]{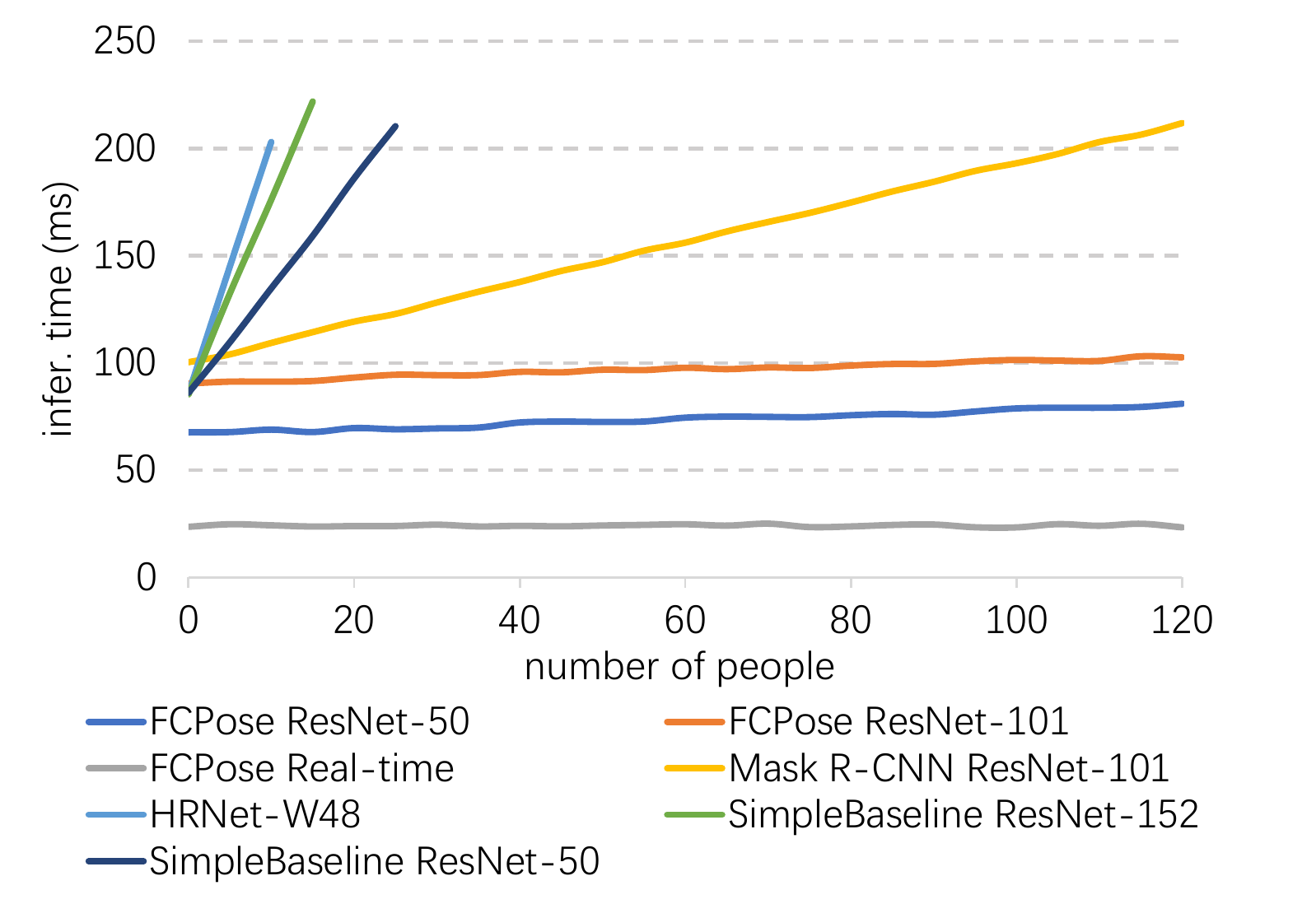}
\caption{
\textbf{End-to-end inference time {w.r.t.}\ the number of the persons in %
an input image.} 
As shown
here, 
the inference time of previous ROI-based methods significantly increases %
in
the number of instances in the image. In sharp contrast, 
the inference time needed for \Ours remains almost constant, which is desirable for real-time applications. 
}
\label{fig:ninst_cropped}
\vspace{-2em}
\end{figure}

In this section, we compare \Ours with other state-of-the-art multi-person pose estimation methods. Unless specified, all the experiments of \Ours in this section use the $3\times$ training schedule, and the performance is reported on the COCO \texttt{test}-\texttt{dev} split.

\noindent\textbf{Comparisons with top-down methods.} As shown in Table~\ref{table:comparisons_with_sota}, compared with the previous top-down method Mask R-CNN, with ResNet-50, \Ours has better performance (64.3\% AP$^{kp}$ vs. 63.9\% AP$^{kp}$). FCPose also has lower computational cost (191.7 GFLOPs \textit{vs.}\ 
212.7 GFLOPs%
) and can infer faster (68ms vs. 89ms per image). If ResNet-101 is used as the backbone, the performance of Mask R-CNN can be only improved from 63.9\% AP$^{kp}$ to 64.3\% AP$^{kp}$ while the performance of \Ours can be boosted from 64.3\% to 65.6\%. We conjecture that the low resolution of ROIs in Mask R-CNN hampers the performance. On the contrary, \Ours\ eliminates the ROIs and thus can bypass the issue.

There are some other top-down methods such as HRNet~\cite{sun2019deep} which first employ an isolated object detector to detect the person box and then crop the ROIs on the original image. The features of these ROIs are computed separately by another network. These methods often have high performance but very slow if we measure the end-to-end inference time (\ie, from the input image to the keypoint results). As shown in Table~\ref{table:comparisons_with_sota}, compared the top-down method HRNet, \Ours can significantly reduce the end-to-end inference time from from 337ms to 68ms (ResNet-50) or 488ms to 93ms (ResNet-101) per image, making the keypoint detection nearly real-time. Additionally, since these ROI-based methods use a relatively cumbersome network to obtain the heatmaps for each ROI \textit{separately}, their total inference time heavily depends on the number of the instances. For example, as shown in Fig.~\ref{fig:ninst_cropped}, the inference time of the model HRNet-W48 significantly increases with the number of the instances. In a sharp contrast, \Ours keeps almost constant inference time. This advantage of \Ours\ is of great importance to real-time applications.

It is also important to note that \Ours does not rely the boxes predicted by the underlying detector. As a result, the keypoint detection results will not be affected by the inaccurate boxes of the detector. In contrast, the ROI-based methods can only predict the keypoints inside the ROIs. If the detector does not yield an accurate box, some keypoints will be missing. As shown in Fig.~\ref{fig:ours_vs_maskrcnn}(left), in Mask R-CNN, the keypoints outside the box are missing. However, as shown in the right figure, \Ours can still correctly detect all the keypoints even if the box is not accurate.

\noindent\textbf{Comparisons with bottom-up methods.} Moreover, we also compare \Ours to bottom-up methods~\cite{newell2017associative, cao2018openpose, kreiss2019pifpaf, papandreou2018personlab}. As shown in Table~\ref{table:comparisons_with_sota}, CMU-Pose~\cite{cao2018openpose} takes 74ms per image to infer and achieves 64.2\% AP$^{kp}$, while \Ours takes 67ms per image and has even better performance (64.2\% AP$^{kp}$). \Ours also achieves better or competitive performance with other bottom-up methods but it can infer much faster.

Finally, some qualitative results are shown in Fig.~\ref{fig:vis_results}, demonstrating that \Ours can work reliably in many challenging cases. These results are based on the ResNet-101 backbone.

\subsection{Real-time Keypoint Detection with \Ours}
\begin{table}[t]
    \small
    \centering
    \setlength{\tabcolsep}{2pt}
	\begin{tabular}{l |c|c|cc|cc}
		method & time (ms) & AP$^{kp}$ & AP$^{kp}_{50}$ & AP$^{kp}_{75}$ & AP$^{kp}_{M}$ & AP$^{kp}_{L}$ \\
		\Xhline{2\arrayrulewidth}
		CMU-Pose~\cite{cao2018openpose} & 74 & 64.2 & 86.2 & 70.1 & \textbf{61.0} & 68.8 \\
		\Ours (DLA-34) & \textbf{24} & 64.8 & 88.4 & 71.4 & 59.6 & 73.3 \\
	    \Ours (DLA-60) & 30 & \textbf{65.9} & \textbf{89.1} & \textbf{72.6} & 60.9 & \textbf{74.1} \\
	\end{tabular}
	\vspace{0.5em}
	\caption{Comparison of the real-time models on the COCO \texttt{test}-\texttt{dev} split. Ours (DLA-34 backbone \cite{yu2018deep} with $736\times 512$ input resize) is more than $3\times$ faster than the previous strong real-time baseline CMU-Pose while %
	obtaining 
	better performance.
	With DLA-60, our performance can be boosted by 1.1\% AP$^{kp}$ while the inference speed is increased by 6ms.
	}
	\label{table:realtime_models}
	\vspace{-2em}
\end{table}
We also present a real-time \Ours\ using 
DLA-34 \cite{yu2018deep} as the backbone. For the real-time model, following FCOS~\cite{tian2019fcos}, the model is trained with $4\times$ training schedule (\ie, 360K iterations). The initial learning is set to 0.01 and it is decayed by a factor of 10 at 300K and 340K, respectively. The shorter side's  size of the input images is reduced from 800 to 512. Moreover, the FPN feature levels $P_6$ and $P_7$ are removed. We also make use of more aggressive data augmentation during training, \ie, the shorter side' size of the input images is sampled from $[128, 736]$.

The performance of the real-time model is shown in Table~\ref{table:realtime_models}. Compared to previous strong real-time baseline CMU-Pose~\cite{cao2018openpose}, our real-time model can run at $\sim$42 FPS on a single 1080Ti GPU and it is 3 times faster (24ms \textit{vs.}\ 
74ms) while having better performance (64.8\% \textit{vs.}\  64.2\% AP$^{kp}$).

\section{Conclusion}
We have proposed a novel keypoint detection framework, termed \Ours. It can eliminate the ROI operations in top-down methods and the grouping post-processing in bottom-up methods, solving keypoint detection in the fully convolutional fashion. The core idea of \Ours is to use the dynamic keypoint head instead of ROIs to make the model attend to instances. Extensive experiments demonstrate that \Ours offers  a simple, fast and effective keypoint detection framework. Additionally, we %
have
presented a real-time version of \Ours that can %
execute 
at $\sim$42 FPS on a single 1080Ti GPU with 64.8\% AP$^{kp}$ on the COCO dataset, outperforming previous strong real-time baseline CMU-Pose~\cite{cao2018openpose} by a large margin.

{\small
\bibliographystyle{ieee_fullname}
\bibliography{draft}
}

\end{document}